\RequirePackage{letltxmacro}
\LetLtxMacro{\LaTeXtextbf}{\textbf}

\documentclass[letterpaper, 10 pt, conference]{ieeeconf}  %
\LetLtxMacro{\textbf}{\LaTeXtextbf}

\IEEEoverridecommandlockouts                              %

\usepackage{cite}

\usepackage[pagewise]{lineno}
\usepackage{float}
\usepackage{graphicx}
\graphicspath{{./figures/}}
\usepackage{multicol}        %
\usepackage[bottom]{footmisc}%
\usepackage{hyperref}
\usepackage[export]{adjustbox}
\usepackage{cite}
\hypersetup{bookmarksopen,bookmarksnumbered,
pdfpagemode=UseOutlines,
colorlinks=true,
linkcolor=blue,
anchorcolor=blue,
citecolor=blue,
filecolor=blue,
menucolor=blue,
urlcolor=blue
}
\usepackage{marvosym}
\usepackage[super]{nth}

\usepackage[T1]{fontenc}    %
\usepackage{upgreek}
\usepackage{amsfonts,amssymb}
\usepackage{bm}
\usepackage{bbm}

\usepackage{amsthm} %
\usepackage{thmtools}
\usepackage{mathtools}
\usepackage{textcomp}
\usepackage{epstopdf}
\usepackage{xspace}
\usepackage{outlines}
\usepackage[]{caption} %
\usepackage[font=footnotesize]{subcaption}
\usepackage{units}
\usepackage{colortbl}
\usepackage{tabulary}
\usepackage[usenames,dvipsnames]{xcolor} %
\usepackage{diagbox}
\usepackage[ruled,vlined,linesnumbered]{algorithm2e}  %
\SetKwComment{Comment}{$\triangleright$\ }{}
\usepackage{ifthen,version}
\usepackage{soul}
\usepackage{fixltx2e}
\usepackage{csquotes}
\usepackage{microtype}      %
\usepackage{lipsum}
\usepackage{csquotes}
\usepackage{tikz}
\let\labelindent\relax
\usepackage{enumitem}
\usepackage{wrapfig}
\usepackage[normalem]{ulem}

\usepackage{multirow}
\usepackage{pbox}
\usepackage{blindtext}
\usepackage{duckuments}
\usepackage{booktabs}
\usepackage{tabularx}
\usepackage{boldline, bbm}
\usepackage{color}
\usepackage[amssymb]{SIunits}
\usepackage{makecell}

\usepackage[ruled]{algorithm2e}

\usepackage[colorinlistoftodos,prependcaption,textsize=tiny]{todonotes}
\usepackage{enumitem}
\setlist[itemize]{noitemsep,nolistsep,leftmargin=17pt}
\setlist[enumerate]{noitemsep,nolistsep,leftmargin=17pt}
\usepackage[font={small}]{caption}
\newcommand{\fref}[1]{Fig.~\ref{#1}} %
\newcommand{\tref}[1]{Table~\ref{#1}} %

\newcommand{\xxnote}[3]{}
\ifx\hidenotes\undefined
  
  \renewcommand{\xxnote}[3]{\color{#2}{#1: #3}}
\fi

\title{\LARGE \bf Learning-on-the-Drive: Self-supervised Adaptive Long-range \\ Perception for High-speed Offroad Driving
}

\author{Eric Chen$^{*1}$, Cherie Ho$^{*2}$, Mukhtar Maulimov$^{2}$, Chen Wang$^{3}$, and Sebastian Scherer$^{2}$%
\thanks{*Both authors contributed equally.}%
\thanks{$^{1}$Eric Chen is with the Department of Computer Science and Mathematics,
        Harvey Mudd College, USA
        {\tt\small erchen@hmc.edu}}%
\thanks{$^{2}$Cherie Ho, Mukhtar Maulimov, and Sebastian Scherer are with the Robotics Institute, Carnegie Mellon University, USA 
        {\tt\small \{cherieh, mmaulimo, basti\}@andrew.cmu.edu}
        }%
\thanks{$^{3}$Chen Wang is with the Spatial AI \& Robotics Lab, University at Buffalo, NY 14260, USA {\tt\small chenw@sairlab.org}}
}%

\begin{document}

\maketitle
\thispagestyle{empty}
\pagestyle{empty}

\begin{abstract}

Autonomous offroad driving is essential for applications like emergency rescue, military operations, and agriculture. Despite progress, systems struggle with high-speed vehicles exceeding $10\meter/\second$ due to the need for accurate long-range ($>50\meter$) perception for safe navigation. Current approaches are limited by sensor constraints; LiDAR-based methods offer precise short-range data but are noisy beyond $30\meter$, while visual models provide dense long-range measurements but falter with unseen scenarios. To overcome these issues, we introduce \textit{ALTER}, a \textit{learning-on-the-drive} perception framework that leverages both sensor types. \textit{ALTER} uses a \textit{self-supervised} visual model to learn and adapt from near-range LiDAR measurements, improving long-range prediction in new environments without manual labeling. It also includes a model selection module for better sensor failure response and adaptability to known environments. Testing in two real-world settings showed on average 43.4\% better traversability prediction than LiDAR-only and 164\% over non-adaptive state-of-the-art (SOTA) visual semantic methods after 45 seconds of online learning. 
\end{abstract}

\section{INTRODUCTION}

Autonomous off-road driving is pivotal in diverse applications including emergency rescue operations \cite{queralta2020collaborative}, military maneuvers \cite{naranjo2016autonomous}, and agriculture \cite{ghobadpour2022off}. These fields demand high-precision and reliable navigation systems capable of operating under varied and challenging conditions. The essence of this requirement lies in the ability to traverse off-road terrains at high velocities, often exceeding $10\meter/\second$, which necessitates long-horizon perception and decision-making capabilities.

\begin{figure}[t]
\centering
  \includegraphics[width=\columnwidth]{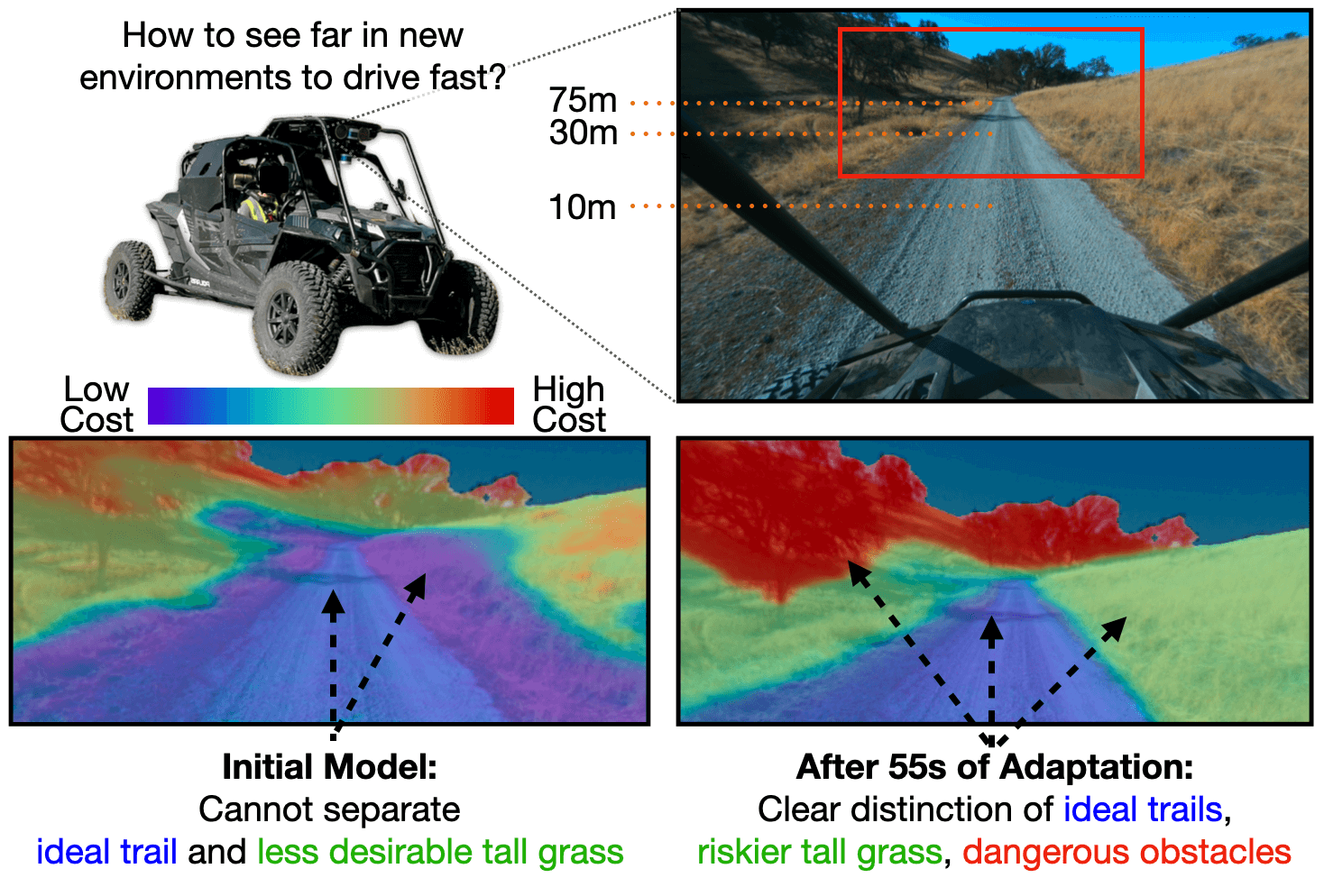}
   \captionof{figure}{We present \textit{ALTER}, a long-range \textit{learning-on-the-drive} perception framework for high-speed offroad driving. We validate it using All-Terrain Vehicle data from two real-world offroad scenarios where key terrains (trails, grass and obstacles) can be differentiated at long range ($>50\meter$) with less than $1~\minute$ of online learning.} 
   \label{fig:main}
\end{figure}

Despite current autonomous offroad driving systems having made significant strides, they continue to grapple with the complexities presented by high-speed vehicular movement. As shown in \fref{fig:main}, the crux of this challenge is the need for accurate long-range perception, specifically beyond $50\meter$. Such capabilities are critical for early decision-making, which is essential for safe and effective navigation in dynamic off-road environments. For example, a robot can operate faster if it distinguishes long-range features like impassable bushes, risky tall grass, and efficient trails \cite{dahlkamp2006self, zhou2012self}.

Most autonomous off-road driving systems are designed for slow speeds below $5\meter/\second$ and tailored for specific scenarios, as evidenced by various studies \cite{Langer_1994_behavioroffroad, Kelly-2004-120769, Bagnell2010learningnavigation, Maturana2017offroad}. 
A key challenge in developing high-speed vehicles is the limitations of sensor technologies.
For example, LiDAR-based perceptual models, though accurate and reliable against visual variations, see reduced effectiveness beyond $30\meter$ due to sparse, noisy data, complicating detailed analysis \cite{Kelly-2004-120769, triest_2023_riskmap}.
In contrast, visual perceptual models offer denser and longer-range observations but rely on extensive human-annotated datasets and falter in unseen scenarios and environments \cite{Maturana2017offroad, Guan2022ganav}.

To effectively utilize the strengths of LiDARs and cameras, an effective way is to self-supervise long-range visual models by near-range LiDAR measurements in an online adaptive setting. For example, traversability can be learned in colored point cloud space \cite{Sofman2006ImprovingRN}, however restricting the prediction to within reliable LiDAR range may limit strategic planning. 
Some methods\cite{dahlkamp2006self,maier2011selfobstacle,zhou2012self} learn from First-Person-View (FPV) images but only yield binary segmentation masks. Such methods are insufficient for high-speed driving which requires \textit{finer-grained terrain reasoning} for early, precise decisions.

\begin{figure*}[ht!]
    \centering
    \includegraphics[width=1\textwidth]{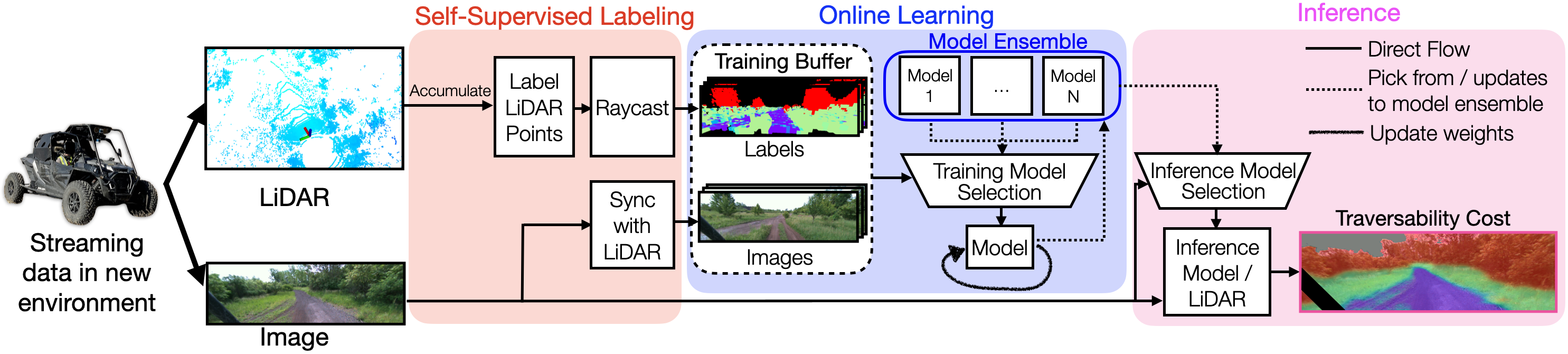}\    \caption{\textbf{\textit{ALTER} Framework:} Our adaptive approach estimates traversability from images using models trained online with new LiDAR data. 
    \textcolor{red}{\textit{Online Labeling}}: 
    We accumulate LiDAR features to label  their costs in 3D. 
    These labels are projected onto the image plane to form training labels. \textcolor{blue}{\textit{Online Learning}}: \textit{ALTER} picks the most similar model or makes a new model for training to adapt to new appearances.
    \textcolor{magenta}{\textit{Online Inference}}: 
    We pick the best inference model given image similarity or LiDAR to predict pixel-wise fine-grained traversability scores. }
    \vspace{-10pt}
    \label{fig:pipeline}
\end{figure*}

To produce fine-grained terrain traversability estimation and achieve online adaptation ($<1~\minute$) simultaneously, we present \textit{\textbf{ALTER}} (\textbf{A}daptive \textbf{L}ong-range \textbf{T}raversibility \textbf{E}stimato\textbf{r}), a novel learning-on-the-drive perception framework that is engineered to be both simple and highly efficient.
It features a straightforward visual model capable of learning from near-range LiDAR measurements in real-time, delivering accurate, pixel-wise estimates.
This involves accumulating near-range LiDAR data to generate a comprehensive long-range map and then projecting these 3D labels onto images for self-supervised learning, leading to a precise, fine-grained, long-range, and online-adaptable perception model.

Additionally, to address potential forgetting issues and manage sensor failures, we introduce a model selection mechanism that dynamically chooses the segmentation head to finetune, creating an ensemble of local experts. This approach provides consistent performance upon re-visiting previously learned environments and yields robust predictions in case of sensor impairments, like cameras obscured by mud, a common and previously overlooked issue in off-road driving.

Our contributions can be summarized as follows:
\begin{itemize}[noitemsep,topsep=0pt,parsep=0pt,partopsep=0pt]
    \item \textbf{Online Learning Framework}:
    We present a simple yet effective online adaptable traversability estimation framework, \textit{ALTER}, based on self-supervised learning. To the best of our knowledge, it is one of the first methods to generate precise, fine-grained, and pixel-wise traversability scores by combining the strengths of LiDAR and camera. Online adaptation is made possible using accumulated LiDAR labels during operation, which also helps to extend the fine-grained perception range from 30\meter~to 100\meter, enabling high-speed off-road driving.
    In addition, we introduce a model selection module employing image-based novelty detection and online validation loss, alleviating the forgetting issue and enhancing robustness against visual sensor failures.
    \item \textbf{Real-World Adaptive Performance}:
    To demonstrate the performance for online adaptation, we evaluate \textit{ALTER} on real-world visually distinct off-road data, which is collected from forests and hills in Pittsburgh, PA, and San Luis Obispo, CA, respectively.
    \textit{ALTER} outperforms LiDAR-only methods on average by $43.4\%$ and non-adaptive methods by $164\%$ with under $1~\minute$ of online learning. Additionally, we show that the model selection module effectively counters catastrophic forgetting and boosts robustness to visual sensor failures.
\end{itemize}

\section{RELATED WORKS}

\textbf{Traversability estimation:}
Traversability estimation involves determining how suitable a specific terrain is for the vehicle to drive over. 
The two most common sensors for offroad perception are cameras and LiDARs. We find works that use LiDARs to quantify traversability by building a grid map whose cells contain geometric statistics \cite{Overbye2022voxel, Langer_1994_behavioroffroad, triest_2023_riskmap}. However, LiDAR measurements are sparse, and therefore noisy, at the distance. Many works incorporate visual predictions to increase the range \cite{Maturana2017offroad, Guan2022ganav, mei2017scene}. For example, Maturana et al. \cite{Maturana2017offroad} classifies an image into semantic classes, such as trail, grass, etc. However, these visual models are trained with copious hand-labeled data and fail when deployed in environments different from the training distribution.

\textbf{Cross-modal Self-supervised Learning:}
To alleviate the need for hard-to-acquire human annotation, self-supervised learning (SSL) is often employed to generate labeled data automatically, with applications in healthcare\cite{tiu2022xrayselfsuper} and robotics\cite{Sivaprakasam2021interact}.
This paradigm has been used to combine complementary information across multiple sensors. For example, visual models have successfully learned from IMU\cite{hdif2023,mayuku2021selfsuper}, force-torque signals \cite{wellhausen2019wheretowalk}, acoustic waveforms \cite{zurn2021acoustic} and odometry \cite{schmid2022reconstruct, yao2022rca,seo2023selfsuper}. Recently, SSL has leveraged exciting progress in visual foundation models like Segment Anything \cite{kirillov2023segany} for improved perception. However, the above works train on static datasets and may still fail when out of distribution.

\begin{figure*}[]
\centering
    \includegraphics[width=\textwidth]{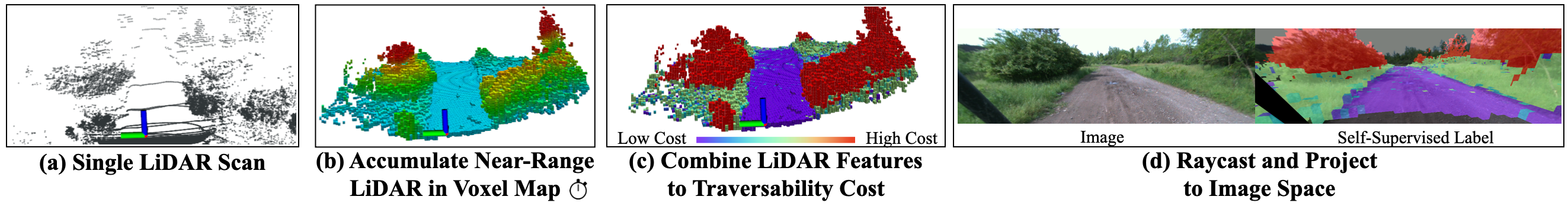}\   
    \caption{\textbf{Self-Supervised Traversability Labeling Pipeline:} 
    Our self-supervised approach generates pixel-wise labels without manual labeling. 
    First, LiDAR scans are registered and accumulated into a voxel map. Then, LiDAR features (object height, terrain slope, planarity) are extracted from the voxel map to calculate traversability costs. Finally, the 3D labels are projected into image space to generate pixel-wise labels. Our labels are able to distinguish between regions of different traversability, such as trail, grass and tall obstacles.}
    \vspace{-10pt}
    \label{fig:training_labels}
\end{figure*}

\textbf{Online Learning:}
Online learning is a way to address data drift by adapting to new data distributions. Common techniques include repeatedly training with the latest training buffer \cite{hadsell2008deep} or discarding outdated clusters of data \cite{dahlkamp2006self}. With larger image datasets, finetuning from pretrained model enables more sample-efficient online learning \cite{Lee2023pretrainedcontinual}. However, fine-tuning a model on new tasks can lead to "forgetting" of previously learned information, known as catastrophic forgetting \cite{FRENCH1999cataforget}. Continual learning approaches prevent forgetting by memory-based \cite{dautume2019episodic} and regularization-based methods \cite{kirkpatrick2017ewc}.

\textbf{Online Self-Supervised Learning for Traversability Estimation:}
For high-speed off-road driving, it is important to have accurate sensing at range, especially in new settings. We find works that leverage cross-modal self-supervised learning and online learning to provide dense visual traversability estimation in unseen environments. Several works train visual traversability models using stereo measurements as supervision\cite{hadsell2008deep,Happold2006terrain}. Hadsell et al. \cite{hadsell2008deep} segments images into multiple classes based on stereo-estimated object height. However, the authors also state the unreliability of stereo range sensing to appearance and lighting changes as a key limitation.
Recent works leverage odometry error to self-supervise visual models online \cite{sathyamoorthy2022terrapn, frey2023fast}. However, they are constrained by the need for direct terrain traversal to learn terrain costs. Frey et al. \cite{frey2023fast} address this by predicting a confidence score to capture uncertain or untraversable regions, but this may lead to initial conservative estimates. Alternatively, LiDAR offers both long-range and untraversable terrain labeling, largely addressing these limitations.

Our work builds on successful approaches that use online self-supervised learning to combine LiDAR and camera \cite{dahlkamp2006self, Bagnell2010learningnavigation, Sofman2006ImprovingRN}. 
Sofman et al. \cite{Sofman2006ImprovingRN} learns a mapping between LiDAR-based traversability to a colored point cloud space. However, the estimation is limited to within reliable LiDAR range while our method goes beyond. We also find works that infer on images \cite{dahlkamp2006self, maier2011selfobstacle, zhou2012self}. However, in these works, the prediction is limited to \textit{binary} traversable or non-traversable classes, which is overly conservative for complex off-road scenes. In contrast, \textit{ALTER} provides \textit{fine-grained} traversability estimation. To address the limitations of past work, we seek to build an adaptive visual model that learns from LiDAR measurements to provide fine-grained, pixel-wise predictions.

\section{APPROACH}

To estimate fine-grained traversability at long-range while being robust to previously unseen environments, we adapt our model online for long-range visual observations by accumulating near-range LiDAR measurements. The framework of \textit{ALTER} is shown in \fref{fig:pipeline}, consisting of three key steps:
\begin{enumerate}[noitemsep,topsep=0pt,parsep=0pt,partopsep=0pt]
    \item Extracting image labels from accumulated LiDAR scans;
    \item Adapting a visual model with the self-labeled data;
    \item Infer on incoming images with online-learned models. 
\end{enumerate}

\subsection{Online Self-Supervised Labeling}
LiDAR can produce accurate near-range traversability labels, hence with the vehicle moving forward, we can obtain the long-range visual labels for the current time step by simply \textit{accumulating} the future near-range LiDAR measurements. This involves (1) pixel-wise labeling in LiDAR space and (2) projecting LiDAR labels to image space.

\subsubsection{Pixel-wise Labeling in LiDAR Space}
To better label LiDAR scans for traversability estimation, we first a) accumulate LiDAR scans into voxel maps; then b) extract the object height, slope, and planarity features; and last c) combine features to inform traversability for off-road driving.

\paragraph{Accumulating LiDAR Scans into Voxel Map}

LiDAR scan accumulation is sensitive to the transform accuracy, hence we first register LiDAR scans with a LiDAR odometry method SuperOdometry \cite{zhao2021-superodometry} as shown in \fref{fig:training_labels}(a).
SuperOdometry is a real-time and robust odometry method, which can meet our expectation of online adaptation.
The resulting registered scans are then accumulated in a $120\meter \times 120\meter$ robot-centered 3D voxel map at a 0.25\meter~resolution (\fref{fig:training_labels}(b)).

\paragraph{Extracting LiDAR Features}
After getting an online voxel map, we extract object height, gravity-aligned ground slope, and planarity features from the accumulated map to inform traversability. Leveraging these features together is valuable for off-road driving \cite{triest_2023_riskmap,lalonde_2006_terrainclass}.
For example, object height indicates tall obstacles, slope helps identify hard-to-traverse terrain, and planarity can differentiate between free space areas such as grass and trail.
For object height $f_{\text{h}}$, we estimate the ground plane with a Markov Random Field-based Filter that interpolates between the heights of the neighboring lowest dense cells. For each voxel in a voxel pillar, we label its object height by the difference between the height of the tallest voxel and ground. The ground plane is also used to derive terrain slope $f_{\text{s}}$.
For planarity $f_{\text{p}}$, we find the eigenvalues for each cell by applying singular value decomposition on the cell's LiDAR points. Using the eigenvalues, we calculate planarity by $\frac{(\lambda_2 - \lambda_3)\times2}{\lambda_1}$. Planarity calculation benefits from accumulation as it requires dense and accurately registered LiDAR data, which is less likely at a distance.

\paragraph{Combining Features to Traversability Costs}
We generate a continuous traversability score by combining LiDAR features into a cost function. This function scores terrain by increasing severity: a trail with low height and high planarity will be low cost, grass with low height and low planarity will be medium cost, regions with high terrain slope is also medium cost, and untraversable obstacles above a height threshold have high costs. 
For efficiency, we simply use the following cost function $J$, where $h_{\text{thresh}} = 1\meter$:
\begin{equation}
   J(f_{\text{h}}, f_{\text{p}}, f_{\text{s}}) = \begin{cases}
        10 & \text{if } f_{\text{h}} \geq h_{\text{thresh}} \\
        6 - f_{\text{p}} + f_{\text{s}}\in[0,6] & \text{if } f_{\text{h}} < h_{\text{thresh}}
    \end{cases} 
\end{equation}
An example labeled 3D voxel map is shown in \fref{fig:training_labels}(c).

\subsubsection{Projecting LiDAR Labels to Image}
Given labeled voxels in 3D space, we project them to image space to generate pixel-wise label masks for visual traversability learning. \fref{fig:training_labels}(d) shows an example image/label pair.

\paragraph{Near-to-far Accumulation}
To mitigate long-range LiDAR noise, we perform a \textit{near-to-far learning} similar to \cite{ Wellington_2006_generativeMRF}, which accumulates data from a short-range sensor to supervise long-range models. Specifically, we associate an image taken at time $t$ with a voxel map after a delay at $t+d$. The resulting labels are more accurate at farther range and cover a larger distance.
To capture accurate labels, we trim the admissible voxels to be within 30\meter~of past and future robot odometries, within which LiDAR provides good estimates, and within $100 \meter$ of the odometry from the image frame.

\paragraph{Raycasting from Camera View}
While projecting a future accumulated map improves the labeling accuracy, a naive projection of LiDAR labels will also include labels that would have been occluded from the initial viewpoint. We use Open3D \cite{Zhou2018Open3d} to render the raycasted view of the voxel labels from the pose where the image is taken. By labeling \textit{accumulated} LiDAR measurements in a near-to-far manner, we can accurately label pixels up to 100m along the driven-over path, instead of just 30m. This results in more accurate and filled-out pixel-wise labels in image space.

\subsection{Online Learning}
As new image/label pairs are generated from self-supervision, we continuously train models to \textit{adapt-on-the-drive} and predict fine-grained pixel-wise traversability scores. 

\begin{figure}[t]
\centering
  \includegraphics[width=\columnwidth]{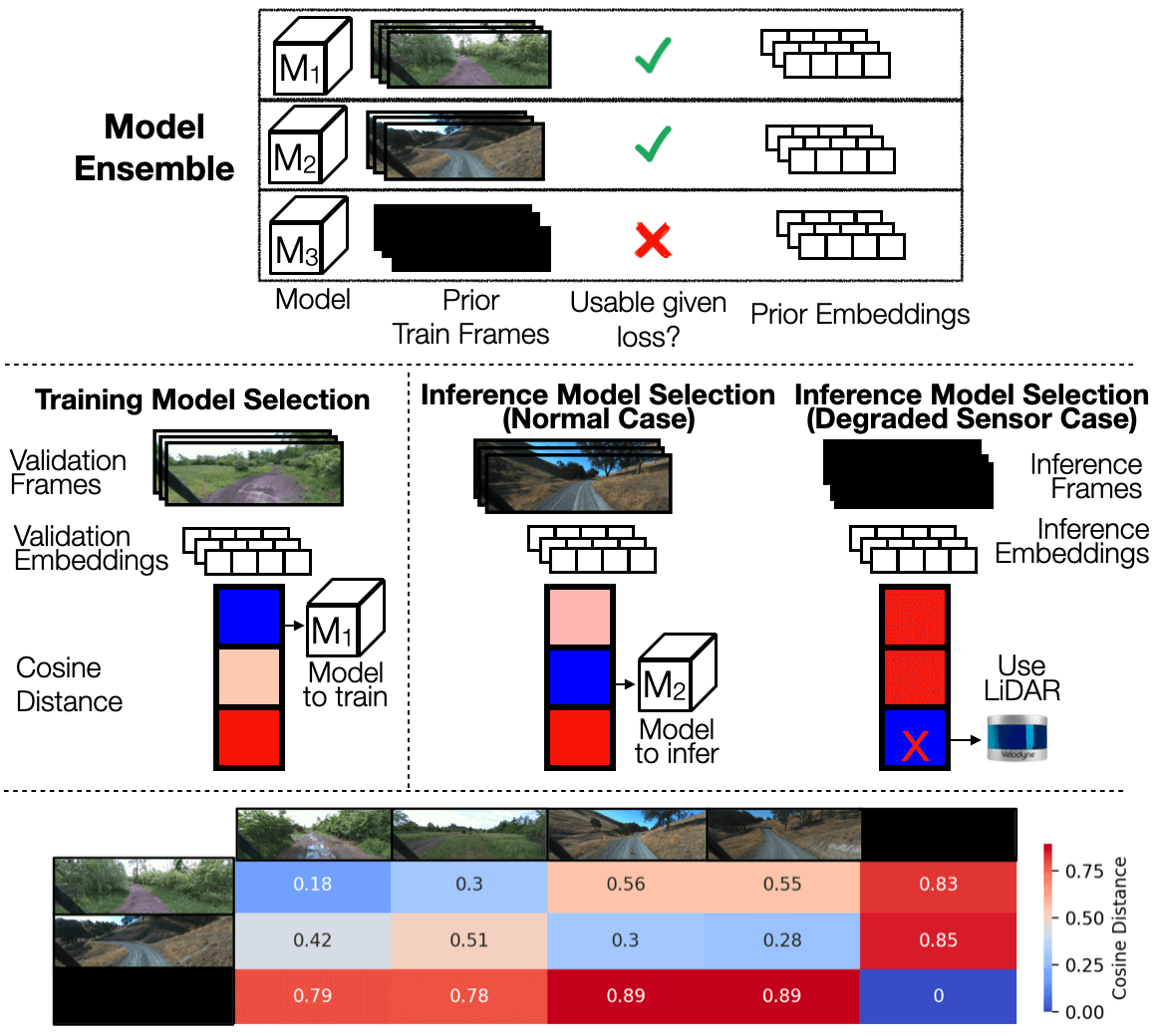}
   \captionof{figure}{\textbf{Training and Inference Model Selection}: For each training cycle, the model with the most similar embeddings to the validation frames is trained. If all models are dissimilar, a new model head is created. The inference model is selected using cosine distance. If this model has not been marked as \textit{usable}, \textit{ALTER} chooses LiDAR. The bottom shows cosine distances between different frames. }  
   \label{fig:model_select}
\end{figure}

Training and inference are done in the top-half of the image as the goal is long-range perception. 
We use a U-Net model \cite{ronneberger2015convolutional} as the base architecture with a MobileNet backbone \cite{howard2017mobilenets} pre-trained on ImageNet \cite{deng2009imagenet} for fast training. The encoder is frozen to keep embeddings aligned during continued training. 
\subsubsection{Training Model Selection}
During operation, the vehicle may visit new and old environments with appearance variations. We therefore keep an ensemble of up to \textit{N} specialized models, each fine-tuned for their settings, allowing for fast reactions when revisiting environments. Our model selection process is similar in spirit to \cite{dahlkamp2006self} with three main differences, (1) their similarity measure is color-based while ours is image-based to include more global context, (2) theirs is purely similarity-based which lacks robustness, and (3) cannot degrade to LiDAR. \fref{fig:model_select} illustrates our \textit{joint loss-similarity model selection} process. At a high level, we combine (1) image feature similarity-based selection which is generally a real-time signal, and (2) loss-based selection which provides an absolute performance metric but lags behind the actual image as labels rely on accumulated LiDAR.
The training model selection starts upon receiving an incoming training buffer. A validation buffer is made from a mix of previous and incoming frames. From the ensemble, we select one candidate to train on the incoming frames by using the shared encoder to generate embeddings and choosing the model whose prior training data embeddings have the lowest cosine difference to the embeddings of the validation frames.
If all models' cosine difference exceeds a threshold $CD_{\text{new}}$, a new model is made by copying the weights of the most similar model. If the model ensemble has reached capacity, the least recently used model is discarded. This strategy is vital for maintaining model relevance while adapting to new environments. 

\begin{figure*}[]
    \centering
    \includegraphics[width=0.9\textwidth]{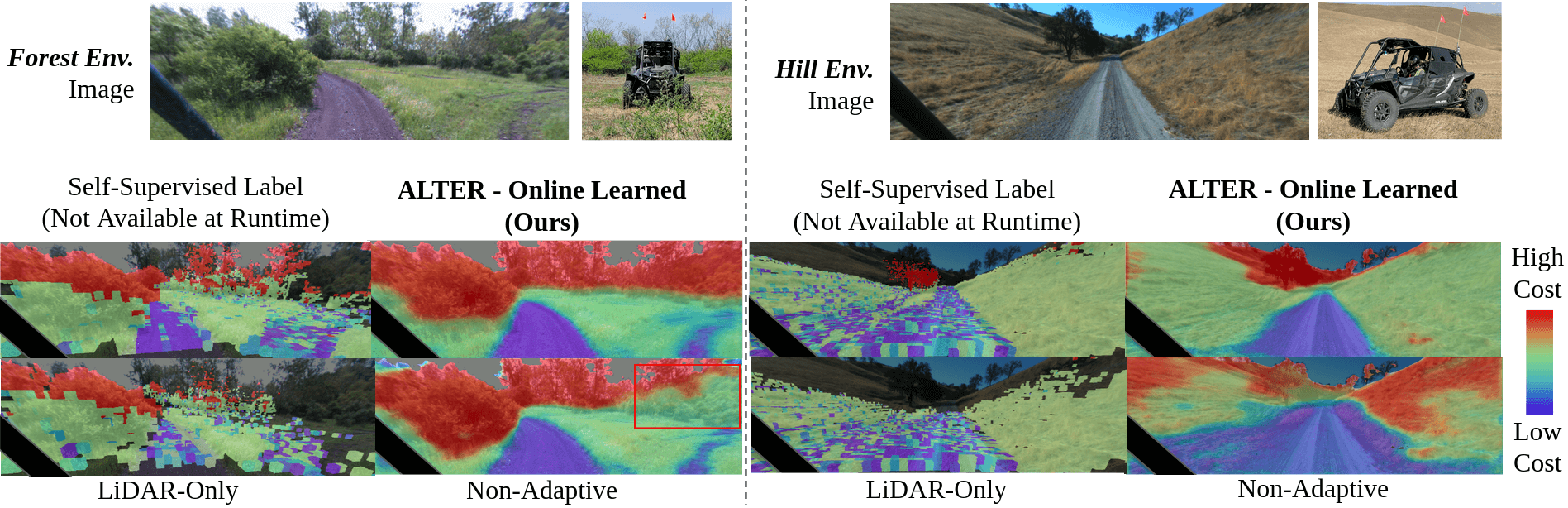}
    \caption{\textbf{Qualitative Continuous Cost Prediction Results after 45s of Adaptation: } 
    Our online-learned method is able to differentiate nuanced terrain (trail, grass, obstacle) at long-range with less than one minute of combined data collection and training. Compared to LiDAR-only baseline, our method can better differentiate grass and trail at a distance, even beyond voxel range. Compared to a non-adaptive model, our method provides more robust predictions when deployed in a new environment. 
    }
    \vspace{-10pt}
    \label{fig:comp_cont}
\end{figure*}

\subsubsection{Model Training} 
To enable fast adaptation, we train the selected model for 10 seconds with a buffer of 10 frames from the most recent experience. Then, the training process restarts using the next selected model and buffer. The training duration is aligned to the buffer creation time so that the next cycle of model selection and training begins as soon as a new buffer of training data is available, improving model reactivity. To mitigate the impact of unknown pixels, the visual model is trained with an adapted Mean Squared Error using only pixels with self-assigned labels. To counter data imbalance, we weight losses on pixels of different traversability ranges using median frequency balancing \cite{Badrinarayanan2017segmnet}.

\subsubsection{Training Model Usability Evaluation}
New model heads are not always preferred due to reasons such as sensor failures, e.g., a simple occlusion of the field of view can corrupt a visual model with fine-tuning. Therefore, we evaluate if the newly trained model head is usable instead of LiDAR after each training cycle. Specifically, if the model's loss on the last $20\second$ of data is below a threshold $L_{\text{usable}}$, then the model will be marked as \textit{usable}. Otherwise, we conservatively opt for LiDAR. We find adding this loss-based component is critical in two settings -- when a model just started learning in a new setting and when a model is training with visually degraded data. LiDAR is the better option in  both scenarios.

\subsection{Inference Method Selection and Traversability Inference}
For efficiency, we select one model head for inference on a per-frame basis, which might not be the model currently training due to training data collection delays.
To be robust to new environments, we compare the last 10 incoming frames' embeddings with those from each model in the ensemble.
If the nearest model is similar enough given a cosine distance threshold $CD_{\text{LiDAR}}$ and has been marked as \textit{usable} given previous loss, the model will be used, otherwise LiDAR will be used. \fref{fig:model_select} shows an example of sensor failure (camera occluded) where LiDAR is used for inference.

\section{RESULTS}

Through our experiments, we aim to answer the  following research questions:
\textbf{Q1}: Does \textit{ALTER} enable more robust performance in new environments than baseline visual methods?
\textbf{Q2}: Does our visual model provide more accurate long-range perception than LiDAR-only estimates?
\textbf{Q3}: How does \textit{ALTER} perform in previously-learned environments?
\textbf{Q4}: How does model selection result in a long-range perception system more robust to visual sensor failures?

\subsection{Experimental Setup}

\subsubsection{Platform and Dataset}
We use a Polaris RZR All-terrain Vehicle (ATV) platform equipped with RGB cameras (Multisense S27) and LiDARs (Velodyne VLP-32C) for data collection.
To demonstrate the generalizability of online adaptation, we select visually distinct environments including \textit{forests} featuring dark-green vegetation, green grass, and dark-brown colored trails in Pittsburgh, PA, and \textit{hills} featuring dry yellow grass with sparse trees in San Luis Obispo, CA. We conduct experiments with two sequences from each environment and confine the duration of each sequence to $75 \second$ to showcase the capability for online adaptation.

\subsubsection{Implementation}
The LiDAR voxel map accumulates data for $5 \second$ for each image frame. Training data are gathered in $10 \second$ buffers, each containing $10$ training and $4$ validation frames. Labels are generated at an average of $0.65 \second$ per frame. After a $10 \second$ buffer is collected, the model trains for $10 \second$ at a learning rate of $3e^{-3}$. Images are captured at $1024 \times 750$ pixels and resized to $512 \times 384$ pixels.
We use $L_{\text{usable}} = 6$, $CD_{\text{new}}=0.45$ and $CD_{\text{LiDAR}}=0.75$.

\begin{table*}[t]
\centering

\caption{\textbf{IOU Comparison After 45s of Adaptation in New Environments}: We compare our adaptive method to baselines on hand-labeled ground truths. \textit{ALTER} is more robust in new environments than the baseline visual model and outperforms all baselines with less than a minute of data collection and adaptation time. (*) Self-supervised labels are not available at runtime as they accumulate future data. They are included for comparison to show that \textit{ALTER} outperforms even its own training data which while informative can be noisy.}
\label{tab:comp_baselines_fullhalf}
\label{tab:method_performance_comparison}
\begin{tabular}{l|@{\hspace{5pt}}r@{\hspace{5pt}}r@{\hspace{5pt}}|@{\hspace{5pt}}r@{\hspace{5pt}}r@{\hspace{5pt}}|@{\hspace{5pt}}r@{\hspace{5pt}}r@{\hspace{5pt}}|@{\hspace{5pt}}r@{\hspace{5pt}}r@{\hspace{5pt}}|@{\hspace{5pt}}r@{\hspace{5pt}}r@{\hspace{5pt}}}
\toprule
\thead{} & \multicolumn{2}{c}{\textbf{Forest Seq. 1}} & \multicolumn{2}{c}{\textbf{Forest Seq. 2}} & \multicolumn{2}{c}{\textbf{Hill Seq. 1}} & \multicolumn{2}{c}{\textbf{Hill Seq. 2}} & \multicolumn{2}{c}{\textbf{Over All Seqs.}} \\ 
 & \textbf{mIoU-10} & \textbf{mIoU-H} & \textbf{mIoU-10} & \textbf{mIoU-H} & \textbf{mIoU-10} & \textbf{mIoU-H} & \textbf{mIoU-10} & \textbf{mIoU-H} & \textbf{mIoU-10} & \textbf{mIoU-H}\\

\midrule
	\textbf{ALTER (Ours)} & \textbf{0.68} & \textbf{0.75} & \textbf{0.62} & \textbf{0.64} & \textbf{0.71} & \textbf{0.71} & \textbf{0.63} & \textbf{0.68} & \textbf{0.66} & \textbf{0.70} \\
Adaptive SVM & 0.35 & 0.42 & 0.28 & 0.34 & 0.55 & 0.59 & 0.45 & 0.52 & 0.41 & 0.47\\
\midrule
ALTER w/o Adapt. & 0.63 & 0.69 & 0.48 & 0.57 & 0.26 & 0.33 & 0.26 & 0.31 & 0.41 & 0.48 \\
GANAV (RELLIS) & 0.32 & 0.35 & 0.29 & 0.26 & 0.11 & 0.13 & 0.11 & 0.13 & 0.21 & 0.22\\
GANAV (RUGD) & 0.48 & 0.45 & 0.46 & 0.45 & 0.12 & 0.13 & 0.13 & 0.18 & 0.30 & 0.30\\
\midrule
Lidar-Only & 0.36 & 0.44 & 0.38 & 0.50 & 0.52 & 0.54 & 0.56 & 0.58 & 0.46 & 0.52\\
\midrule
Self-supervised Label* & 0.54 & 0.59 & 0.55 & 0.61 & 0.59 & 0.63 & 0.61 & 0.63 & 0.57 & 0.62\\
\bottomrule
\end{tabular}

\end{table*}

\begin{figure*}[t]
    \centering
    \includegraphics[width=0.9\textwidth]{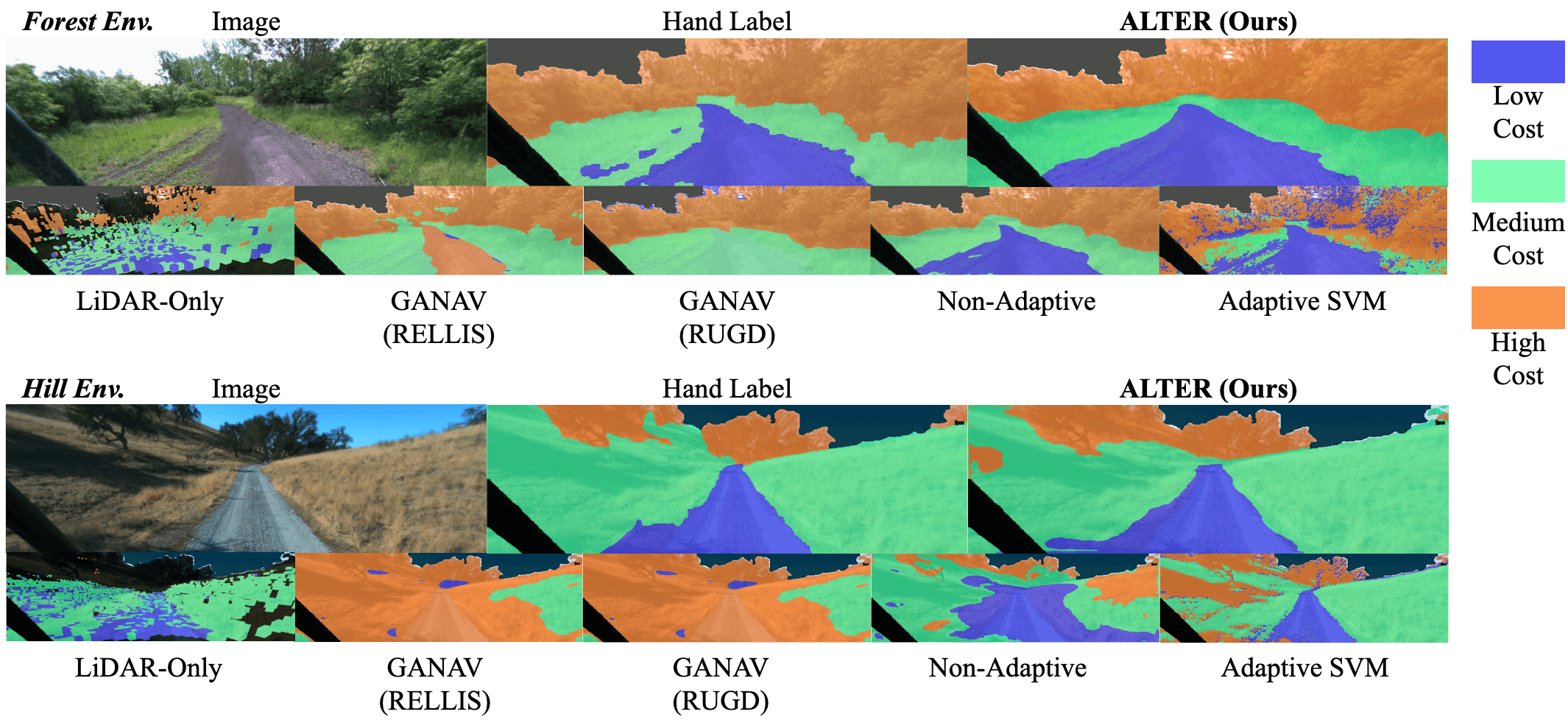}
    \caption{\textbf{Qualitative Comparison to Hand-labels after 45s of Adaptation: } \textit{ALTER} enables more accurate differentiation of the classes at long-range. Compared to non-adaptive methods, our adaptive method provides more robust predictions when deployed in a new environment. }
    \vspace{-10pt}
    \label{fig:comp_seg}
\end{figure*}

\subsubsection{Metrics}

We provide intersection-over-union (IOU) metrics using two comparisons, (1) \textit{mIoU-10}: Pixels beyond 10 meters where all methods have labels, and (2) \textit{mIoU-H}: full top-half image (removing sky and robot cage) to include areas beyond LiDAR, using bilinear interpolation to fill LiDAR gaps. Metrics are calculated against images manually segmented into three classes (low, medium, and high cost/untraversable) based on an expert operator's sense of terrain desirability. In general, low cost reflects smooth trails, medium cost reflects grassy or steep areas, and high-cost reflects untraversable obstacles or dangerous cliffs. 
For comparison, we convert continuous cost predictions to discrete classes and map semantic class predictions from GANAV \cite{Guan2022ganav} to these three classes, as common in semantic navigation\cite{Maturana2017offroad}. For \textit{ALTER}, we threshold 0-2.5 to low, 2.5-7.5 to medium and 7.5 to high-cost. For GANAV, we map \textit{smooth, bumpy} to low, \textit{rough} to medium and \textit{obstacle, forbidden} to high-cost.

\subsubsection{Baselines}
We select four baselines for comparison at long range and in out-of-distribution environments.

\begin{enumerate}[noitemsep,topsep=0pt,parsep=0pt,partopsep=0pt]
    \item \textbf{LiDAR-only}: The geometric-based traversability map built from accumulated LiDAR points up to current time is commonly used for offroad driving \cite{triest_2023_riskmap,Kelly-2004-120769}. For comparison, the 3D scores are raycasted into image space. This differs from self-supervised labels, which incorporate future data and are not available in real-time.
    
    \item \textbf{GANAV\cite{Guan2022ganav} (Trained on RELLIS dataset\cite{jiang2021rellis}}): State-of-the-art visual semantic model for offroad driving, with author-provided weights pretrained on RELLIS dataset. 
    \item \textbf{GANAV\cite{Guan2022ganav} (Trained on RUGD dataset \cite{RUGD2019IROS})}: As above, with author-provided weights for RUGD dataset.
    \item \textbf{Adaptive SVM, based on \cite{zhou2012self}:} This classic method incorporated a self-supervised adaptive visual model, the closest method to ours. It adapts a Support Vector Machine (SVM) online given LiDAR labels. For efficiency, we implement a version with Linear SVM and Stochastic Gradient Descent. 
    As it predicts binary labels, we modify the approach to output three classes for fair comparison. 
\end{enumerate}
While there has been recent work on adaptive self-supervised visual traversability estimation using odometry \cite{frey2023fast, sathyamoorthy2022terrapn}, the unavailability of source code and use of a different supervision source render a direct comparison inappropriate.

\begin{table}[t]
\caption{\textbf{Comparison when Adapting to Multiple Domains}: Adaptive methods (top rows) outperform non-adaptive methods (mid rows) on sequences with multiple environments. \textit{ALTER}'s model selection helps it to quickly adapt, leading to improved performance within the first 15 seconds (F15) of revisiting a similar environment.}
\label{tab:mixed_env}
\centering
\resizebox{\linewidth}{!}{
\begin{tabular}{l|cccc|cccc}
\toprule
\thead{} & \multicolumn{4}{c}{\textbf{Pretrained $\rightarrow$ Forest 2 $\rightarrow$ Hill 2}} & \multicolumn{4}{c}{\textbf{Pretrained $\rightarrow$ Hill 2 $\rightarrow$ Forest 2}}\\ 
 \textbf{mIoU} & \textbf{10} & \textbf{H} & \textbf{10-F15} & \textbf{H-F15} & \textbf{10} & \textbf{H} & \textbf{10-F15} & \textbf{H-F15}\\
\midrule
\textbf{ALTER (Ours)} & \textbf{0.69} & \textbf{0.67} & \textbf{0.64} & \textbf{0.66} & \textbf{0.70} & \textbf{0.68} & \textbf{0.64} & \textbf{0.66}\\
ALTER w/o MS & 0.60 & 0.63 & 0.39 & 0.52 & 0.65 & 0.65 & 0.51 & 0.58\\
Adaptive SVM & 0.35 & 0.39 & 0.16 & 0.18 & 0.44 & 0.46 & 0.39 & 0.36\\

\midrule
GANAV (RELLIS) & 0.21 & 0.22 & 0.23 & 0.26 & 0.20 & 0.22 & 0.23 & 0.26\\
GANAV (RUGD) & 0.25 & 0.27 & 0.18 & 0.24 & 0.24 & 0.26 & 0.18 & 0.24\\
\midrule
LiDAR-Only & 0.48 & 0.52 & 0.44 & 0.49 & 0.48 & 0.52 & 0.44 & 0.49\\
\bottomrule
\end{tabular}
}
\end{table}

\subsection{Does ALTER enable robust prediction in new scenarios?}
First, we test \textit{ALTER}'s robustness in new environments. 
We test on four sequences total across two environments, with the first $45\second$ for training and the next $30\second$ for evaluation. \textit{ALTER} starts with a model trained on both sequences of the opposing environment.

\tref{tab:comp_baselines_fullhalf} quantitatively compares \textit{ALTER} with the baselines, showing \textit{ALTER} provides more accurate long-range results across all sequences, with less than 1 minute of adaptation each. Adaptive SVM struggles to distinguish grass and trees of similar colors due to its reliance on color and texture features. Our U-Net-based approach leverages larger contextual information and attain more precise estimates. 
To isolate the impact of online learning, we compare to \textit{ALTER without adaptation}, and show an average of 61\% improvement. Importantly, we found that GANAV \cite{Guan2022ganav}, a SOTA visual semantic model trained on open datasets, fails in our data, suggesting the need for online adaptation. \fref{fig:comp_cont} shows representative qualitative outputs and \fref{fig:comp_seg} shows comparisons to hand-labeled classes. \textit{ALTER} clearly distinguishes low-cost planar trails, medium-cost grassy and sloped regions, and high-cost obstacles in both environments. In contrast, the non-adaptive methods fail to predict the trail accurately, as shown in the Hill Env. 
This result shows that \textit{ALTER} improves robustness in new settings.

\begin{table}[t]
\caption{\textbf{Comparison over a Mission with Simulated Sensor Failure}: \textit{ALTER} outperforms all baselines while \textit{ALTER w/o MS} and other vision-based methods show poor resilience.}
\label{tab:sensor_failure}
\centering
\resizebox{\linewidth}{!}{
\begin{tabular}{l|@{\hspace{5pt}}r@{\hspace{5pt}}r@{\hspace{5pt}}r@{\hspace{5pt}}r@{\hspace{5pt}}|@{\hspace{5pt}}r@{\hspace{5pt}}r@{\hspace{5pt}}r@{\hspace{5pt}}r@{\hspace{5pt}}}
\toprule
\thead{} & \multicolumn{4}{c}{\textbf{Forest 2 w/ Failure}} & \multicolumn{4}{c}{\textbf{Hill 2 w/ Failure}} \\ 
 \textbf{mIoU} & \textbf{10} & \textbf{H} & \textbf{10-F15} & \textbf{H-F15} & \textbf{10} & \textbf{H} & \textbf{10-F15} & \textbf{H-F15}\\
\midrule
\textbf{ALTER (Ours)} & \textbf{0.52} & \textbf{0.59} & \textbf{0.52} & \textbf{0.60} & \textbf{0.65} & \textbf{0.61} & \textbf{0.68} & \textbf{0.60} \\
ALTER w/o MS & 0.30 & 0.34 & 0.26 & 0.32 & 0.38 & 0.35 & 0.36 & 0.34 \\
Adaptive SVM & 0.27 & 0.31 & 0.30 & 0.35 & 0.27 & 0.30 & 0.34 & 0.38\\
\midrule
GANAV (RELLIS) & 0.26 & 0.29 & 0.25 & 0.27 & 0.16 & 0.14 & 0.15 & 0.16 \\
GANAV (RUGD) & 0.42 & 0.29 & 0.43 & 0.32 & 0.13 & 0.20 & 0.12 & 0.16 \\
\midrule LiDAR-Only & 0.40 & 0.50 & 0.38 & 0.48 & 0.55 & 0.53 & 0.52 & 0.48 \\
\bottomrule
\end{tabular}
}
\end{table}

\subsection{Does ALTER provide longer-range estimates?} 
This section examines \textit{ALTER}'s ability to determine traversability at long-range compared to LiDAR. We observe that \textit{ALTER} can provide $43\%$ improvement in traversability estimates over LiDAR at long range.
In the first row of \fref{fig:comp_seg}, we show a scenario where our online-learned prediction differentiates the trail from the grass region at long range. In contrast, the LiDAR-only map does not distinguish the two areas well because the planarity feature of the trail requires a high density of points to be accurate. Moreover, in the second row of \fref{fig:comp_seg}, the LiDAR-only baseline is unable to provide estimates at long range due to the small voxel map range (60m radius) needed for fast runtime. In contrast, \textit{ALTER} can predict the trail and grass much further than even the voxel map size. As observed, \textit{ALTER} is able to provide better long-range traversability estimates than using only LiDAR.

\begin{figure}[t]
\centering
  \includegraphics[width=\columnwidth]{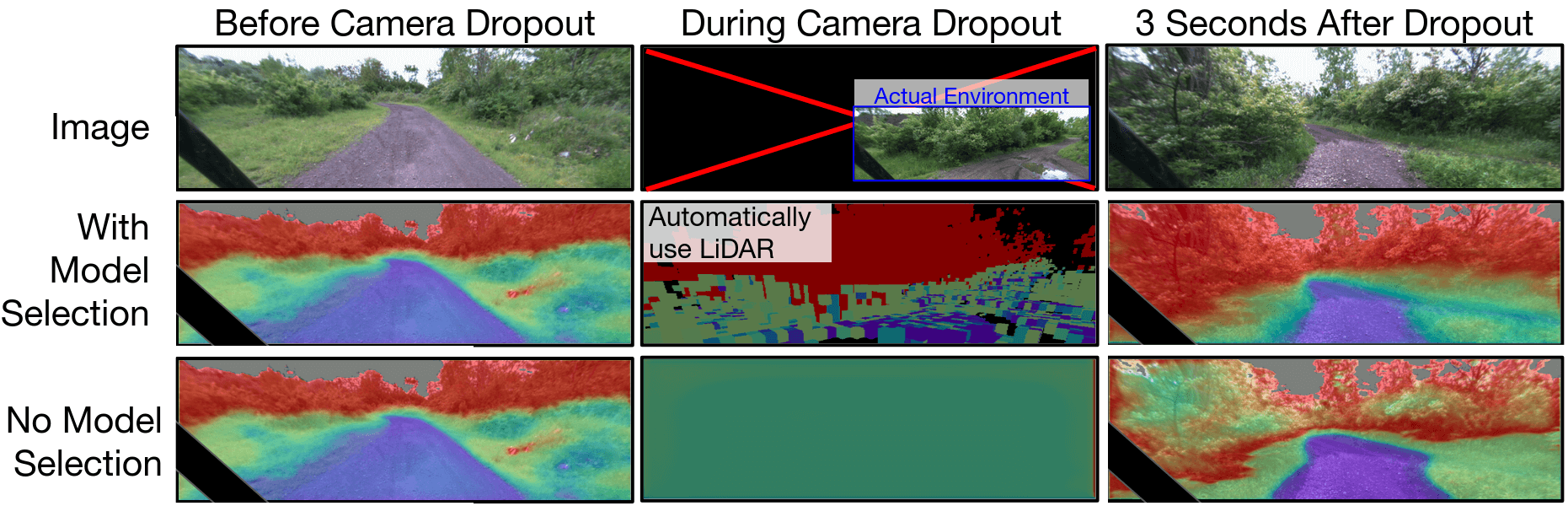}
   \captionof{figure}{\textbf{Camera Dropout Experiment}: Without model selection, the system predicts erroneously using the visual model during dropout. In contrast, \textit{ALTER} automatically uses LiDAR during camera dropout and quickly recovers when the sensor is back.}  
   \label{fig:dropout}
\end{figure}

\subsection{How does \textit{ALTER} perform in learned environments?}
We also validate \textit{ALTER}'s ability to overcome catastrophic forgetting \cite{FRENCH1999cataforget} with a multiple-scenario test setup. We provide two starting models for our system, one model trained on Seq. 1 of each environment. We test on two scenarios, Forest$\rightarrow$Hill, Hill$\rightarrow$Forest, where the vehicle transitions from one starting environment to another. To isolate the benefit of model selection, we also compare to \textit{ALTER without model selection (ALTER w/o MS)}. For adaptive methods without model selection mechanism, we average the results from runs beginning from each starting model.
In addition, we report the mIoU of the first 15s in each environment, denoted \textit{F15}, to quantify a method's reactiveness to a new environment.

Table \ref{tab:mixed_env} compares the different methods, with \textit{ALTER} consistently outperforming baselines. With model selection, \textit{ALTER} significantly improves its reactiveness to new environment with much higher accuracy in the first 15 seconds than \textit{ALTER without model selection}. While slower to react, we find even without model selection, \textit{ALTER} also outperforms all other baselines, giving credit to the adaptive nature.

\subsection{How does model selection operate for visual degradation?}

We next examine how \textit{ALTER}'s model selection module leads to more robustness against visual sensing failures. We follow the same experimental setup as the previous subsection. We test with sudden simulated \textit{View Drop}, a standard camera failure tested in Computer Vision communities \cite{ge2023metabev}. We simulate camera dropout of various lengths (between 20s and 50s) in each environment, to test each method's ability in times without camera images and when camera images come back. \tref{tab:sensor_failure} and \fref{fig:dropout} display the comparison, where \textit{ALTER} outperforms all methods. \textit{ALTER without model selection} performs worse than LiDAR-only method, which is expected as LiDAR provides consistent, albeit shorter-range measurements during visual sensing failure. In addition, without model selection, the method continues to fine-tune on the faulty sensor data and cannot rapidly recover when the camera is normal. 
In contrast, \textit{ALTER} can switch immediately to LiDAR when it senses it is out-of-distribution and then immediately uses the visual model when the sensor gives in-distribution inputs. This experiment validates that the model selection module can robustify against visual impairment.

\section{CONCLUSIONS}
In this paper, we present \textit{ALTER}, an offroad perception framework that \textit{adapts on-the-drive} to predict dense traversability map in novel environments. \textit{ALTER} is a self-supervised framework that \textit{adapts} a visual model online using near-range LiDAR measurements. We incorporate a model selection module to pick the best model to train and, at inference, pick the best model for predictions or use LiDAR instead. We validate with real-world offroad driving datasets that \textit{ALTER} significantly increases robustness when operating in new environments. We showed that the adapted visual model can provide more accurate long-range measurements than LiDAR, especially on terrains that require dense points to differentiate. Finally, we show that \textit{ALTER} provides more reactivity when revisiting previously learned environments and improved robustness against visual sensing degradation. For future work, we are interested in incorporating visual foundation models \cite{kirillov2023segany, jung2023v} for faster adaptation in harder-to-segment images, pretraining on large multi-modal datasets \cite{sivaprakasam2024tartandrive2, cheng2023treescope}, incorporating more data sources such as proprioception \cite{hdif2023} for more robustness against LiDAR noise, and more intelligent data sampling \cite{wang2021unsupervised} for improved performance.

\section*{ACKNOWLEDGMENT}

Distribution Statement `A' (Approved for Public Release, Distribution Unlimited). 
This research was sponsored by DARPA (\#HR001121C0189).
The views,
opinions, and/or findings expressed are those of the author(s) and should not be interpreted as representing the
official views or policies of the Department of Defense or the U.S. Government. We thank JSV and EF for their assistance.

{\small
\bibliographystyle{IEEEtran}
\bibliography{./IEEEfull,refs}
}
\end{document}